\documentclass{IOS-Book-Article}

\usepackage{mathptmx}
\usepackage{soul}\setuldepth{article}
\usepackage{booktabs}
\usepackage{multirow}
\usepackage{listings}
\usepackage{graphicx}
\usepackage{comment}
\usepackage[numbers]{natbib}

\def\hb{\hbox to 11.5 cm{}}

\begin{document}

\pagestyle{empty}
\def\thepage{}
\begin{frontmatter}

\title{Unlocking Criminal Court Decisions: Flexible Information Extraction for Empirical Legal Research}

\author[B]{\fnms{Tomáš} \snm{KNAP}\orcid{0009-0009-2085-9182}%
\thanks{Corresponding Author: Tomáš Knap, tomas.knap@prf.cuni.cz.}},
\author[A]{\fnms{Ivana} \snm{KVAPILÍKOVÁ}\orcid{0000-0003-1479-3294}},
\author[B]{\fnms{Jan} \snm{ČERNÝ}\orcid{0009-0006-4873-8233}},
\author[A]{\fnms{Klára} \snm{BENDOVÁ}\orcid{0000-0001-8002-6566}},
\author[C]{\fnms{Jaromír} \snm{ŠAVELKA}\orcid{0000-0002-3674-5456}},
\author[B]{\fnms{Vojtěch} \snm{POUR}\orcid{0009-0005-4075-6579}} and
\author[B]{\fnms{Jakub} \snm{DRÁPAL}\orcid{0000-0001-9455-9013}}

\runningauthor{T. Knap et al.}
\address[A]{Faculty of Mathematics and Physics, Charles University, Prague, Czechia }
\address[B]{Faculty of Law, Charles University, Prague, Czechia}
\address[C]{School of Computer Science, Carnegie Mellon University, Pittsburgh PA, USA}

\begin{abstract}
Criminal verdicts contain descriptions of criminal behavior that can support quantitative legal research, but structuring them requires attributes and extraction rules that cannot always be fully specified in advance. We present an LLM-based pipeline that builds a structured dataset through attribute exploration, text span extraction, and value normalization, guided by annotation instructions that legal experts can refine. We evaluate it on two traffic offenses: driving under the influence (DUI) and driving despite a ban (DDB). For DUI, revising the instructions was associated with higher extraction accuracy (from 52\% to 70\% of correctly extracted values) and higher agreement between human annotators. Repeated runs of the same model agreed with each other more than with the gold annotations, showing that consistent output is not necessarily correct, while disagreement between annotators and between models helped reveal unclear instructions. For DDB, where the instructions defined fixed categories and output formats, agreement was high for most attributes.
\end{abstract}

\end{frontmatter}
\thispagestyle{empty}

\maketitle

\section{Introduction}

Court verdicts contain a wealth of information about criminal behavior and its consequences, but much of this information remains locked in unstructured text. Specifically, a detailed description of the criminal behavior forms part of continental European criminal judgments. Since key criminal justice issues rely on understanding what criminal behavior offenders actually commit, analyzing it at scale could fundamentally expand how scholars, practitioners and public institutions study criminal justice. 
Understanding patterns in how courts process cases is further impossible without detailed structured data, making it a crucial step for better functioning of sentencing councils, ministries of justice and everyday work of criminal lawyers.

Until recently, collection of these data was tedious and impossible on a large scale as it required individual collection of court decisions and human classification of descriptions of criminal behaviors contained in these verdicts. At present, many European countries publish full-text court verdicts \citep{drapal_digitalised}, making large-scale analyses feasible. The challenge is to enable transformation of unstructured verdicts into structured legal data on a large scale for legal experts, who know little about computational methods, but have strong expert knowledge in criminal law. 

\begin{figure}[t]
    \centering
    \includegraphics[width=0.85\linewidth]{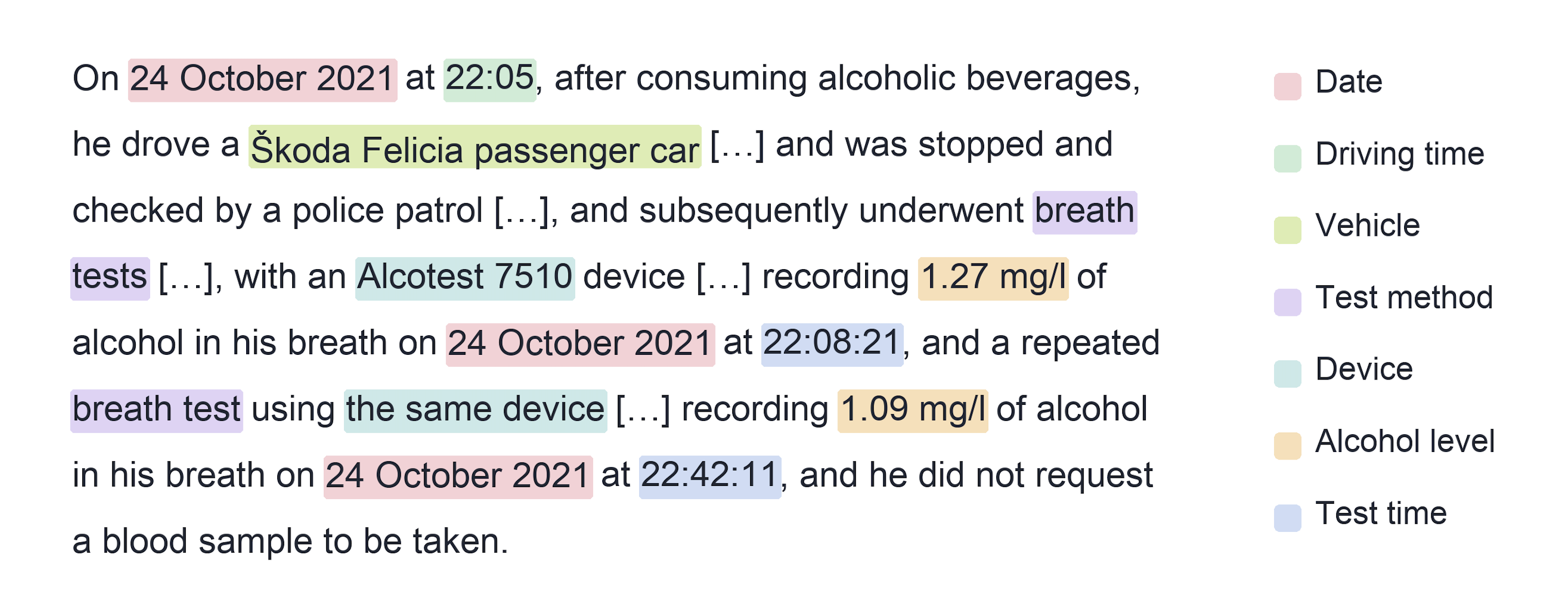}
    \caption{English translation of a shortened factual sentence
    with selected attributes highlighted. The passages are edited for illustration purposes and conciseness (i.e., [\ldots]).}
    \label{fig:factual-sentence-example}
\end{figure}

We investigate how factual descriptions of criminal behavior (Figure~\ref{fig:factual-sentence-example}) can be transformed into structured data for empirical legal research. We ask: How can LLM-assisted exploration support the development of extraction schemas (RQ1)? How does refining annotation instructions relate to annotation consistency and extraction accuracy (RQ2)? What can agreement and disagreement among human annotators and LLMs reveal about the reliability of the resulting data (RQ3)?

Building on prior work \cite{kvapilikova-etal-2025-legal}, we contribute an LLM-driven workflow connecting attribute exploration, evidence-linked extraction, and normalization. We evaluate schema induction and instruction refinement, and analyze disagreements across two offense categories to identify ambiguities and show why repeatability alone does not establish correctness.

\section{Related Work}
Constructing datasets from criminal court decisions often requires \emph{discovering attributes and categories from the data}. Drápal et~al.~\cite{drapal2023thematic} used GPT-4 to support thematic analysis of theft descriptions, finding that legal experts could improve the initial codes through natural language feedback. Gray et~al.~\cite{gray2024factors} generated factors and their definitions from court opinions, obtaining representations that supported outcome prediction but did not match the performance of expert-defined factors. More recently, Ali et~al.~\cite{ali-etal-2026-argumentation} discovered factors from legal documents and LLM knowledge for use in insurance dispute resolution. These studies concern the discovery of themes or legally relevant factors. We investigate the discovery of attributes describing criminal behavior, including dates, measurements, and circumstances whose values must subsequently be extracted and normalized.

There have also been several approaches to \emph{extracting structured information from legal texts}. Janatian et~al.~\cite{janatian2023structure} used LLMs to convert legislation into pathways for legal expert systems. Costa et~al.~\cite{costa2024brazilian} developed a semantic annotation pipeline for Brazilian judicial decisions, representing its outputs as ontology-aligned nanopublications with provenance information. Closer to our task, Bendová et~al.~\cite{bendova-etal-2025-facts} investigated extracting court-established factual descriptions from Slovak criminal opinions. Kvapilíková et~al.~\cite{kvapilikova-etal-2025-legal} then combined expert-defined attributes with data-driven attribute discovery to structure descriptions of traffic-related offenses. The latter study is the direct starting point for our work. We extend its evaluation by examining how revisions to the annotation instructions affect human agreement and automatic extraction across two offense categories.

Our use of annotation instructions to guide extraction is also related to \emph{human--LLM collaboration in qualitative coding}. Dai et~al.~\cite{dai-etal-2023-llm} proposed an iterative collaboration framework for developing and applying a codebook, obtaining coding quality similar to human coders in two survey datasets. Gao et~al.~\cite{gao2024collabcoder} developed CollabCoder to support independent coding, discussion of disagreements, and collaborative codebook development with LLM assistance. Together with the legal thematic analysis framework of Drápal et~al.~\cite{drapal2023thematic}, these studies already establish a role for researchers in directing LLM-assisted coding. Our focus is on specifying extraction rules, cardinalities, and value formats for structured legal datasets, with annotation instructions serving as the interface through which researchers revise these decisions.

The reliability of the resulting data depends in part on \emph{annotation instructions and their refinement}. Schepers et~al.~\cite{schepers2025annotation} found that GPT-4o matched human performance on basic annotations of human-rights decisions but had difficulties with citation recall and exhibited variation across repeated prompts. Savelka et~al.~\cite{savelka2023gpt4} and Kim et~al.~\cite{kim-etal-2026-refining} used discrepancies between LLM annotations and reference labels to refine guidelines for legal sentence classification and biomedical named entity recognition, respectively, reporting modest improvements over the original guidelines.
We study the role of instruction refinement in constructing criminal-behavior datasets, comparing human agreement, repeated runs, cross-model agreement, and agreement with gold annotations. 

\section{Dataset}

We use Slovak criminal court judgments from 2018 to 2022, collected and linked to administrative records in earlier work \cite{bendova-etal-2025-facts}. The administrative dataset contains 126,795 judgments and provides a reference for assessing the coverage of the available full texts, which varies across courts. For the experiments reported here, we randomly sampled 200 factual descriptions concerning driving under the influence (DUI) and another 200 concerning driving despite an imposed driving ban (DDB).

Each criminal behavior description (\textit{factual sentence}) is a passage in the judgment describing the conduct for which the defendant was found guilty. We obtained these passages using a combination of regular expressions and LLM prompting, as described in earlier work \cite{bendova-etal-2025-facts}. A single factual sentence may contain several events and measurements. Figure~\ref{fig:factual-sentence-example} illustrates a driving event followed by two breath tests. Structuring this information requires distinguishing the time of driving from the test times and associating each measured value with the corresponding test.

To identify cases that can be described using a common set of attributes, we group judgments by the conduct described in their factual sentences. Often, high-level legal classifications are insufficient. Driving under the influence of alcohol and operating construction machinery under the influence of methamphetamine fall under the same penal-code provision but involve different behaviors. We therefore used GPT-5 mini to classify factual sentences into conduct-based categories \cite{kvapilikova-etal-2025-legal}. This classification provided the basis for selecting the DUI and DDB samples used in our experiments. Errors in these upstream steps may affect the composition of the resulting dataset, as discussed in \cite{bendova-etal-2025-facts}, but not the evaluation reported here.

We further restrict the experimental dataset to single-offense judgments, which account for 68\% of the 126,795 judgments and which we identify from the sentencing provisions cited in the operative part of the judgment. Multiple and continuing offenses, whose factual sentences describe several acts, are left for future work. 

\section{Pipeline Description}

\begin{figure}
    \centering
    \includegraphics[width=\textwidth]{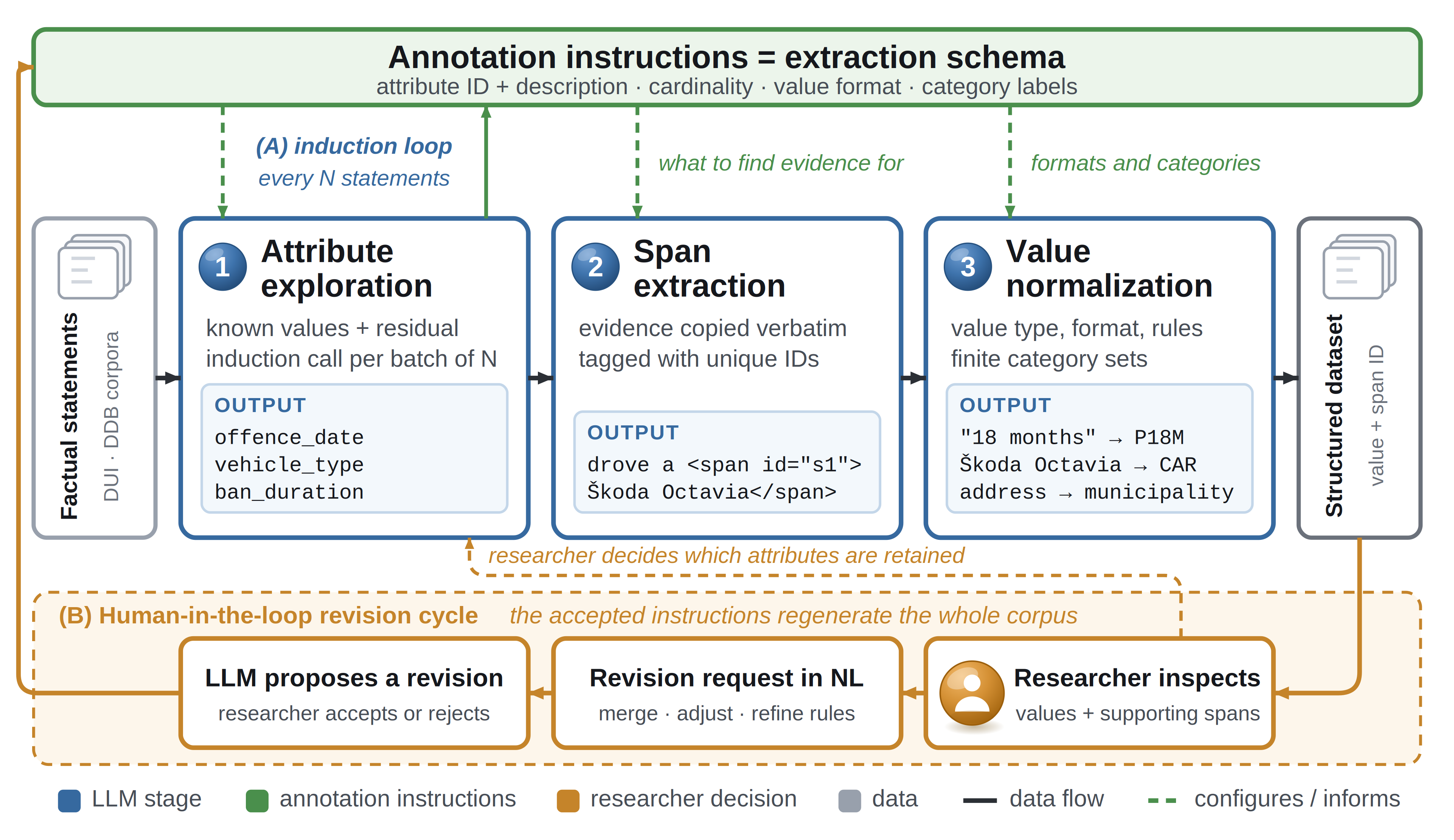}
    \caption{The extraction pipeline. The annotation instructions serve as the extraction
schema for all three stages: (1) attribute exploration, (2) span extraction and (3)
value normalization. Loop (A) induces new attributes every $N$ statements;
cycle (B) is the researcher's revision of the annotation instructions.}
    \label{fig:attex_diagram}
\end{figure}

Our pipeline supports iterative development of an extraction schema and normalization rules through corpus exploration.\footnote{Code and supplementary materials: \href{https://github.com/tomasjknap/attex-pipeline}{github.com/tomasjknap/attex-pipeline}.} As shown in Figure~\ref{fig:attex_diagram}, it consists of three stages: (1) attribute schema induction (\emph{attribute exploration}), (2) extraction of supporting text spans (\emph{span extraction}), and (3) normalization to a canonical format or class label (\emph{value normalization}). The annotation instructions specify each attribute's identifier, description, cardinality, value format and, where relevant, category labels. The induction loop (A) proposes additional attributes, while the researcher revision cycle (B) refines the instructions based on analytical objectives and extracted data (Section~\ref{sec:human-oversight}).

\subsection{Attribute Exploration}
\label{ssec:schema-induction}

Attribute exploration (stage 1 in Figure~\ref{fig:attex_diagram}) discovers recurring concepts without requiring a complete predefined schema. Building on an LLM-based pipeline \cite{kvapilikova-etal-2025-legal}, statements are processed in fixed-size batches against the current schema. After every $N$ statements, residual information not covered by existing attributes is passed to an LLM, which proposes candidate attributes for subsequent batches. Accepted candidates are merged into the schema. An expert may provide an initial schema, or the process begins with an induction step. For each statement, one LLM call extracts values for existing attributes, and another identifies information outside the current schema. One additional call per batch uses this residual information to propose new attributes.

The main difficulty is controlling semantic granularity. Equivalent concepts may receive different names across runs, while superficially similar expressions may denote distinct concepts. For example, a DUI statement may contain separate dates for the offense, a breath or blood test, and a prior judgment. Frequency alone is also insufficient for filtering candidates because rare but genuine attributes must be distinguished from incidental one-off details.

We enforce non-overlap through explicit instructions in the induction prompt, repeated at several points to discourage the model from proposing attributes that overlap with existing concepts. Batch size $N$ provides an additional control over granularity. Small batches tend to produce overly specific attributes, while large batches increase context demands and may overlook rare but relevant information.

\subsection{Attribute Extraction: Span Extraction}
\label{ssec:attr_extraction}

Once attributes are identified, their supporting evidence is located in the original statements (stage 2 in Figure~\ref{fig:attex_diagram}). This makes extracted values traceable to their source and preserves their original expression before normalization.

Evidence can be represented as copied text, character offsets, or identifiers inserted into the statement \citep{goyal-etal-2026-semantic}. The optimal representation varies across models and datasets \citep{ravikumar-etal-2026-lost}, with tagging tending to favor precision and span-based approaches recall \citep{gu-etal-2022-empirical}. Copied text is simple but ambiguous when expressions repeat, while LLM-generated character offsets are sensitive to whitespace, punctuation, and indexing differences \citep{akomeah-etal-2025-quabynar}. Inline tags provide stable references but complicate overlapping, discontinuous, or nested spans.

We therefore first extract evidence as copied text, align it with the original statement, and wrap the aligned spans in generic XML-like tags with unique identifiers. Subsequent stages reference these identifiers:

\begin{verbatim}
The defendant drove a <span id="s1">Škoda Octavia</span>
on <span id="s2">15 March 2024</span> with a blood alcohol
concentration of <span id="s3">1.24 ‰</span>.
\end{verbatim}

\subsection{Attribute Extraction: Value Normalization}

Value normalization (stage 3 in Figure~\ref{fig:attex_diagram}) derives consistent value types, output formats, normalization rules, and, where appropriate, categories from observed values. It also assigns each attribute either single or multiple cardinality, distinguishing attributes with a single value per statement (e.g., \textsc{dateOfOffense}) from list-valued attributes that may have multiple instances (e.g., substances or measurements). For each attribute, the LLM receives the extracted spans with summary statistics on value frequencies and observed cardinality, and uses this evidence to propose the corresponding schema specifications. These choices depend on the expressions encountered in the corpus. For example, driving-ban durations may be expressed in months, years, or implicitly through effective and expiration dates. Locations may be reduced from full addresses to municipalities. Vehicle descriptions may be mapped to categories such as \textit{CAR} or \textit{MOTORCYCLE}.  

Our strategy is to preserve as much information as possible while removing superficial details. This limits transformations that cannot be verified against the supporting evidence and favors rules that can be applied consistently across instances. The resulting normalization rules become part of the schema and are applied during final extraction. Researchers may modify formats and categories or introduce their own categorization schemes.

\subsection{Human Oversight}
\label{sec:human-oversight}

The revision cycle (B in Figure~\ref{fig:attex_diagram}) lets researchers align extraction with their analytical objectives through changes to the annotation instructions. Where statements support multiple interpretations, researchers determine which attributes matter, how values are categorized, and how ambiguities are resolved. Researchers can remove or merge attributes, adjust categories, and refine extraction rules after inspecting extracted values and supporting evidence.

Revisions can be requested in natural language, with an LLM proposing updated instructions for researcher review and acceptance. Corrections to individual extractions can also inform revisions to general rules. Once accepted, the revised instructions can be used to regenerate annotations consistently across the dataset.

\section{Experimental Design}

We evaluate two main components of the pipeline. First, we investigate attribute exploration (Section~\ref{ssec:schema-induction}) on the DUI dataset, focusing on how induction parameters and constraints affect the resulting schema. This experiment studies the induction procedure itself rather than comparing schemas across domains. Second, we evaluate attribute extraction on both DUI and DDB, using the second dataset to examine whether the observed extraction behavior generalizes to a different schema and annotation setting. We consider extraction accuracy as well as agreement across human annotators, two repeated LLM runs, and different LLMs. In our LLM experiments, we use minimal reasoning effort and the default sampling parameters (temperature = 1, top\_p = 1).

To obtain reference annotations, four law students in their second to fifth year of study annotated the sampled factual sentences in INCEpTION software across three rounds. Sentences were randomly assigned, with each independently annotated by two students to measure inter-annotator agreement. Disagreements were resolved into curated reference annotations (gold) by a team member or a trained law student not involved in the original annotation, who saw only the human annotations and no model outputs. The procedure evolved based on interim experimental results and differed between the two datasets. The curation procedure also incorporated refinements informed by our previous work \citep{kvapilikova-etal-2025-legal}, particularly concerning annotation consistency and completeness.

For DUI, annotators selected text spans and assigned predefined attributes and categories in two rounds using successive versions of the annotation instructions. For DDB, annotators answered predefined questions in specified formats, and their responses were converted into structured reference values.

For extraction evaluation, legal researchers proposed and iteratively refined the schema to reflect their research objectives and the observed data. Each specification was expressed as annotation instructions for human annotators and a JSON schema for the extraction pipeline, aligning the definitions used in human and automatic extraction.

\section{Results \& Discussion}

\subsection{Controlling Schema Induction}

A central design choice in attribute exploration is how much evidence the model should observe before proposing new attributes. We first investigated this on a 40-statement subset of the DUI dataset. Four induction rounds over approximately 10 residuals each produced 83 attributes, 17 of which were never populated in subsequent statements. In contrast, a single induction round over all 40 residuals produced 44 attributes with substantially fewer non-recurring attributes. Smaller induction batches therefore encourage more granular, often case-specific proposals.

We observed the same pattern on the full 200-statement dataset. With induction batch sizes of 100 and 200, the resulting schemas contained 39 and 30 attributes, respectively (Table~\ref{tab:population-rates-summary}). With batch size 100, 33\% of attributes were populated in fewer than 5\% of statements, compared with 13\% for batch size 200.

\begin{table}[t]
\centering
\renewcommand{\arraystretch}{1.0}
\caption{Attribute population rates for induction batch sizes 100 and 200 across the same 200 factual statements. Population rate denotes the percentage of sentences in which a value for the attribute was extracted.}
\label{tab:population-rates-summary}
\begin{tabular}{lrr}
\toprule
& Batch size 100 & Batch size 200 \\
\midrule
Attributes                    & 39   & 30   \\
Mean population rate (\%)     & 32.6 & 45.2 \\
Median population rate (\%)   & 10.4 & 48.0 \\
Attributes $<$5\% population  & 13 (33\%) & 4 (13\%) \\
Attributes $<$10\% population & 18 (46\%) & 6 (20\%) \\
\bottomrule
\end{tabular}
\end{table}

Low population alone is insufficient for pruning the schema. Rare attributes such as \textsc{activeDrivingBan} (3.5\%) and \textsc{attemptedBribeOrCorruption} (0.5\%) may still be relevant to quantitative legal research. Retaining an attribute therefore requires considering both its frequency and its relevance to the research question.

A second challenge is semantic duplication and overlap between induced attributes. We tested embedding-similarity deduplication of attribute descriptions but could not identify a similarity threshold that reliably removed duplicate concepts without also merging distinct ones. Applying a prompt-level non-overlap constraint reduced the number of proposed attributes by 18\% in a comparable run.

These observations support treating the induced schema as a set of candidates for researcher review. Batch size and prompt-level constraints help control the proposals, while researchers determine which distinctions are useful for the intended analysis.

\subsection{Schema Specification Determines Extraction Quality}

Once attributes have been identified, how precisely must they be specified for consistent extraction? We examine this on DUI by comparing two extraction schemas, including their attribute definitions, categories, and extraction rules. Schema 1 represents the initial specification, while Schema 2 incorporates revisions informed by the initial annotation and extraction results.

Human inter-annotator agreement indicates how much ambiguity remains in each specification. Two annotators independently annotated the same statements in each round, and agreement on the seven categorized attributes was measured by Cohen's $\kappa$.

\begin{table}[t]
\centering
\renewcommand{\arraystretch}{1.0}
\small
\setlength{\tabcolsep}{2.5pt}
\caption{Agreement on DUI annotations. Human agreement is reported separately for Schema 1 (initial) and Schema 2 (revised), using observed agreement ($P_o$) and Cohen's $\kappa$. All LLM comparisons use Schema 2 and report $\kappa$: between two repeated runs of \texttt{gpt-5.6-terra} (Internal), between this model and GPT-5 mini or GPT-5, and between this model and gold annotations.}
\label{tab:iaa}
\label{tab:kappa-all}
\begin{tabular}{@{}lrrrrrrrr@{}}
\toprule
& \multicolumn{2}{c}{Human: Schema 1}
& \multicolumn{2}{c}{Human: Schema 2}
& \multicolumn{4}{c}{LLM: Schema 2} \\
\cmidrule(lr){2-3}\cmidrule(lr){4-5}\cmidrule(lr){6-9}
Attribute & $P_o$ & $\kappa$ & $P_o$ & $\kappa$
& Internal & vs. Mini & vs. GPT-5 & vs. gold \\
\midrule
SubstanceUsed & 0.714 & 0.38 & 0.859 & 0.64 & 0.86 & 0.40 & 0.78 & 0.84 \\
MeasuringMethod & 0.662 & 0.30 & 0.792 & 0.52 & 0.79 & 0.58 & 0.56 & 0.55 \\
LicenseStatus & 0.179 & $-$0.03 & 0.742 & 0.57 & 0.90 & 0.45 & 0.41 & 0.51 \\
ReasonForCheck & 0.223 & 0.01 & 0.595 & 0.47 & 0.67 & 0.17 & 0.20 & 0.12 \\
RefusedTesting & 0.464 & 0.04 & 0.556 & 0.29 & 0.49 & 0.48 & 0.16 & 0.29 \\
TypeOfRoad & 0.406 & 0.23 & 0.714 & 0.63 & 0.88 & 0.40 & 0.60 & 0.69 \\
VehicleType & 0.803 & 0.39 & 0.899 & 0.60 & 0.92 & 0.81 & 0.79 & 0.77 \\
\bottomrule
\end{tabular}
\end{table}

Human agreement was higher under Schema 2 for all seven categorized attributes (Table~\ref{tab:iaa}). Six reached moderate to substantial agreement ($\kappa=0.47$--$0.64$), while \textsc{RefusedTesting} remained at fair agreement ($\kappa=0.29$), with identical responses in 55\% of cases. These results suggest that refining overall instructions, attribute definitions and categorization rules can support more consistent annotation, although increased annotator familiarity with the task between rounds may also have contributed.

We next examined whether the revised specification was also associated with improved automatic extraction. We ran \texttt{gpt-5.6-terra} over the same 200 DUI factual sentences using each schema and compared the outputs with the corresponding gold annotations.

\begin{table}[t]
\centering
\renewcommand{\arraystretch}{1.0}
\small
\setlength{\tabcolsep}{2.5pt}
\caption{Automatic extraction accuracy against curated reference annotations using the initial and revised schemas. Span accuracy applies to all attributes; value accuracy is reported for the seven categorized attributes. Overall accuracy is the share of attribute values that match the reference annotations across all sentences.}
\label{tab:llm-gold}
\begin{tabular}{@{}lrrrr@{\hspace{1.5em}}lrrrr@{}}
\toprule
& \multicolumn{2}{c}{Schema 1} & \multicolumn{2}{c}{Schema 2}
& & \multicolumn{2}{c}{Schema 1} & \multicolumn{2}{c}{Schema 2} \\
\cmidrule(lr){2-3}\cmidrule(lr){4-5}\cmidrule(lr){7-8}\cmidrule(lr){9-10}
Attribute & span & value & span & value &
Attribute & span & value & span & value \\
\midrule
dateOfOffense   & 0.960 & --    & 0.985 & --    & measuringMethod     & 0.453 & 0.557 & 0.583 & 0.639 \\
timeOfOffense   & 0.874 & --    & 0.975 & --    & substanceUsed       & 0.316 & 0.399 & 0.639 & 0.923 \\
measuringTime   & 0.785 & --    & 0.939 & --    & typeOfRoad          & 0.354 & 0.385 & 0.653 & 0.762 \\
measuringDevice & 0.748 & --    & 0.824 & --    & refusedTesting      & 0.421 & 0.507 & 0.722 & 0.667 \\
amountUsed      & 0.719 & --    & 0.750 & --    & licenseStatus       & 0.125 & 0.282 & 0.345 & 0.655 \\
vehicleType     & 0.758 & 0.909 & 0.789 & 0.945 & reasonForCheck      & 0.112 & 0.217 & 0.098 & 0.156 \\
vehicleModel    & 0.505 & --    & 0.556 & --    & placeDetails        & 0.156 & --    & 0.107 & --    \\
placeOfOffense  & 0.597 & --    & 0.721 & --    & geographicalDetails & --    & --    & 0.352 & --    \\
checkedBy       & 0.485 & --    & 0.512 & --    \\
\cmidrule(l){6-10}
&&&&& \textbf{OVERALL} & \textbf{0.552} & \textbf{0.518} & \textbf{0.639} & \textbf{0.696} \\
\bottomrule
\end{tabular}
\end{table}

We evaluate extraction at two levels: \textit{span accuracy} measures whether the extracted span overlaps with the gold span by at least 50\%, while \textit{value accuracy} measures whether the extracted value is assigned to the correct category or, for formattable attributes, whether it exactly matches the gold value. Overall span accuracy increases from $0.552$ to $0.639$, while value accuracy increases from $0.518$ to $0.696$. The largest improvements occur for attributes whose interpretation was clarified substantially during revision. For \textsc{SubstanceUsed}, value accuracy increases from $0.399$ to $0.923$, and for \textsc{LicenseStatus} from $0.282$ to $0.655$.

\subsection{Agreement as a Diagnostic for Schema Refinement}

Accuracy against gold annotations alone does not reveal why an extraction fails. We therefore compare several forms of agreement on the same categorized attributes: agreement between human annotators, repeated runs of the same LLM, different LLMs, and the primary LLM against gold annotations (Table~\ref{tab:kappa-all}).

Repeated executions of \texttt{gpt-5.6-terra} generally show higher agreement than comparisons with other models or gold annotations. This indicates that repeatability alone does not establish extraction correctness. Cross-model disagreement can help identify cases for review, but may reflect either ambiguous instructions or model errors.

The pattern for \textsc{ReasonForCheck} illustrates this distinction. The attribute requires contextual classification rather than extraction from a direct lexical cue. Human annotators disagree over whether erratic driving indicates \texttt{suspected\_intoxication} or a \texttt{traffic\_violation}, and different models likewise diverge. Agreement is consequently low across human, cross-model, and model-to-gold comparisons. Such disagreement indicates that the category itself requires a more explicit decision rule.

Other attributes require inference beyond what is explicitly stated. For example, \textsc{LicenseStatus} may require deciding whether the absence of evidence about a revoked or missing license implies that the license is valid. Similarly, \textsc{RefusedTesting} may be implicitly false when breath measurement information is provided. Such rules cannot be resolved merely by enumerating category labels. They require the schema to specify what inferences the extractor is permitted to make.

These results motivate the revision cycle (B in Figure~\ref{fig:attex_diagram}). Disagreement helps prioritize expert review: researchers inspect evidence, distinguish extraction errors from unclear instructions, and revise the instructions before repeating extraction.

\subsection{Generalization to Driving Despite a Ban}

The DDB experiment tests whether lessons from the DUI experiments transfer to a second extraction setting. Its extraction design incorporates two constraints motivated by our earlier observations. First, occurrence multiplicity is specified explicitly: ten of the eleven attributes are single-valued, and only \textsc{AdditionalTrafficViolation} is multi-valued. Second, extraction is formulated as question answering with closed categories or explicitly defined output formats instead of open span selection.

\begin{table}[t]
\centering
\renewcommand{\arraystretch}{1.0}
\small
\caption{{Cohen's $\kappa$ for DDB across human, repeated-run,
cross-model, and model-to-gold comparisons.}}
\label{tab:kappa-DDB}
\begin{tabular}{lrrrrr}
\toprule
& Human & LLM & Cross-model & Cross-model & LLM vs.\\
Attribute & IAA & internal & (vs.\ mini) & (vs.\ gpt-5.2) & gold \\
\midrule
DateOfOffense              & 0.789 & 1.000 & 1.000 & 1.000 & 0.980 \\
TimeOfOffense              & 0.848 & 0.990 & 0.985 & 0.980 & 0.965 \\
PlaceOfOffense             & 0.336 & 0.789 & 0.403 & 0.739 & 0.482 \\
OffenseMunicipality        & 0.563 & 0.964 & 0.857 & 0.928 & 0.758 \\
VehicleType                & 0.348 & 1.000 & 0.822 & 0.959 & 0.876 \\
AdditionalTrafficViolation & 0.100 & 0.604 & 0.020 & 0.411 & 0.083 \\
IssuingAuthorityType       & 0.694 & 1.000 & 0.938 & 0.989 & 0.937 \\
IssuingAuthorityLocation   & 0.756 & 1.000 & 0.941 & 0.995 & 0.946 \\
DecisionIssueDate          & 0.684 & 0.989 & 0.939 & 0.995 & 0.920 \\
DecisionEffectiveDate      & 0.691 & 1.000 & 0.944 & 0.995 & 0.919 \\
DrivingBanDuration         & 0.844 & 0.946 & 0.807 & 0.850 & 0.946 \\
\bottomrule
\end{tabular}
\end{table}

Agreement is substantially higher than for DUI across most DDB attributes. Nine of the eleven attributes reach $\kappa \geq 0.68$ in all or nearly all comparisons, with model agreement frequently above $0.9$. This is consistent with the hypothesis that explicit cardinality, closed categories, and constrained output formats limit divergent interpretations.

The remaining disagreements are also informative. \textsc{AdditionalTrafficViolation} is difficult primarily because positive instances are rare and are frequently omitted rather than assigned to competing categories. Of 21 human mismatches, 18 result from one annotator identifying a violation that the other omits. A similar pattern appears in automatic extraction. \textsc{PlaceOfOffense} and \textsc{VehicleType}, in contrast, exhibit relatively low human agreement despite high model agreement. Inspection shows that annotators differ in how they handle partially anonymized sentences when the requested value itself remains explicit, exposing a specific gap in the annotation instructions.

\section{Conclusion} 

We presented an LLM-driven pipeline for transforming descriptions of criminal behavior into structured data linked to supporting text spans. Experiments showed that attribute exploration produced candidate attributes whose granularity depended on induction settings, while their relevance required expert judgment (RQ1). Expert refinement of the schema and extraction instructions was associated with higher human agreement and extraction accuracy on DUI (RQ2). Across both offense categories, disagreements between humans and LLMs exposed unclear definitions and extraction errors, while consistency across repeated LLM runs did not guarantee agreement with gold annotations (RQ3).

These findings support treating dataset creation as an iterative process. Domain experts provide analytical objectives and knowledge that cannot be derived from the data alone, while inspection of extraction errors and disagreements reveals where definitions and rules need clarification. The pipeline connects schema induction, extraction, evaluation, and expert refinement, supporting both verification of the resulting data and revision of the specification used to produce them.

\section*{Acknowledgments} We acknowledge funding by the European Union (ERC StG RE-CON-SENT, project no. 101219946); views and opinions expressed are however those of the author(s) only and do not necessarily reflect those of the European Union or the European Research Council. Neither the European Union nor the granting authority can be held responsible for them. 

OpenAI Codex and Claude Code were used to assist with software development, translation of the factual sentence in Figure~\ref{fig:factual-sentence-example} into English, and preparation of the HTML and XML code used to render Figures~\ref{fig:factual-sentence-example} and~\ref{fig:attex_diagram}. The authors reviewed the resulting code, translation, and figures and take responsibility for their accuracy.

\bibliographystyle{vancouver}
\bibliography{references}

\end{document}